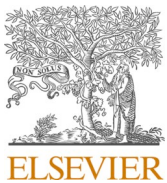
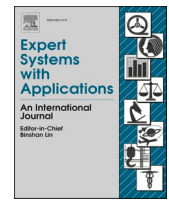

# SkillNER: Mining and mapping soft skills from any text

Silvia Fareri [a,*,1], Nicola Melluso [b], Filippo Chiarello [b], Gualtiero Fantoni [c]

[a] Department of Economics, University of Modena and Reggio Emilia, Italy
[b] Department of Energy Systems, Territory and Construction Engineering, University of Pisa, Italy
[c] Department of Civil and Industrial Engineering, University of Pisa, Italy

ABSTRACT

In today's digital world, there is an increasing focus on soft skills. On the one hand, they facilitate innovation at companies, but on the other, they are unlikely to be automated soon. Researchers struggle with accurately approaching quantitatively the study of soft skills due to the lack of data-driven methods to retrieve them. This limits the possibility for psychologists and HR managers to understand the relation between humans and digitalisation.

This paper presents SkillNER, a novel data-driven method for automatically extracting soft skills from text. It is a named entity recognition (NER) system trained with a support vector machine (SVM) on a corpus of more than 5000 scientific papers. We developed this system by measuring the performance of our approach against different training models and validating the results together with a team of psychologists. Finally, SkillNER was tested in a real-world case study using the job descriptions of ESCO (European Skill/Competence Qualification and Occupation) as textual source.

The system enabled the detection of communities of job profiles based on their shared soft skills and communities of soft skills based on their shared job profiles. This case study demonstrates that the tool can automatically retrieve soft skills from a large corpus in an efficient way, proving useful for firms, institutions, and workers.

The tool is open and available online to foster quantitative methods for the study of soft skills.

## 1. Introduction

In recent years, academia and industry have shown a growing interest in soft skills. The focus of scholars and practitioners on this topic has sharpened for many reasons.

First, the definition of the concept remains vague. Some researchers define soft skills as a set of embedded characteristics of personality traits (Blake & Gutierrez, 2011; Deming & Kahn, 2018). Others consider soft skills as a synergy of multiple competences that could be acquired through experience and knowledge (Robles, 2012; Evenson, 1999; Schulz, 2008; Mitchell et al., 2010). Thus, there still exists the need to settle on a common definition of the concept.

Second, while digitalisation improves businesses and industrial processes (Melluso et al., 2020), digital technologies threaten many classes of workers (Frank et al., 2019; Bridgstock, 2011; Cooper & Tang, 2010). The fear of robotisation seems to be one of the main drivers of the increasing focus on soft skills due to the seminal work of Frey and Osborne (2017), who estimate that around 47% of jobs are at high-risk of robotisation, especially those characterised by routine tasks. The results of this research look encouraging for "job profiles characterised by soft-content related tasks." Consequently, many studies outline the importance of acquiring soft skills in preparation for this digital wave (Weber, 2016; Ummatqul Qizi, 2020; Fareri et al., 2020).

There are several methodological approaches to studying soft skills, epistemic research and qualitative exploration being the main ones. Psychologists and human resources (HR) professionals are familiar with these approaches, which demonstrate promising results: we now have innovative ways to facilitate soft skills development (Sanz et al., 2019; Tseng et al., 2019; Duran-Novoa et al., 2011), assessment (Bohlouli, et al., 2017), comprehension (Chechurin & Borgianni, 2016), or the identification of their impact on the workforce (Hendon et al., 2017).

Moreover, researchers in the field have also adopted methodological approaches that rely on recent advances in information retrieval. The wide availability of data has enabled the adoption of techniques that accelerate the extraction of new insights in the fuzzy domain of soft skills as well. For example, recent natural language processing (NLP) improvements have proven to be suitable for several applications in labour market studies (Fareri et al., 2020). One of the most effective NLP techniques is named entity recognition (NER).

NER is a computational linguistic method capable of extracting and classifying named entities mentioned in unstructured text into pre-defined categories (such as person names, locations, and product names). Assigning a word to a semantic class provides crucial information for tasks such as question answering (Abujabal et al., 2018; Blanco-Fernández et al., 2020), topic disambiguation (Fernández et al., 2012) or detection (Krasnashchok & Jouili, 2018; Lo et al., 2017; Al-Nabki et al., 2019), and revealment of relationships among elements (Sarica et al., 2020; Amal et al., 2019). Furthermore, NER has proved to be effective in broader applications, such as user profiling (Nicoletti et al., 2013) and ontology development in unconventional domains (Oliva et al., 2019; Rodrigues et al., 2019). Recent advancements in artificial intelligence, such as the introduction of transformer-based language models (Devlin et al., 2018)18, has improved dramatically the performances of such systems. However, these systems are now challenged to retrieve uncommon entities, and the NLP community is working hard to make improvements in this direction (Hu et al., 2020). This is the case of soft skills.

The aim of this study is to develop a methodology that helps researchers and practitioners to study soft skills leveraging NER. We developed a tool called SkillNER to extract soft skills from any text.

Furthermore, we offer a demonstration of how SkillNER can help in the study of soft skills. We applied our tool to the database provided by the European Skill/Competence Qualification and Occupation (ESCO) framework. This application led to exploring the labour market from a novel perspective, identifying communities of job profiles and communities of soft skills.

To sum up, we contribute to the literature in three ways. First, we developed SkillNER, an NLP system able to extract soft skills – a novel, vaguely defined, and rare entity. To promote the use of our approach by other scholars in the study of soft skills, SkillNER is also made available as a web application[2].

Second, since SkillNER was built with the help of a supervised system, we developed training data labelled by a panel of domain experts (psychologists). Third, we contribute to the progress in understanding this complex domain by demonstrating the utility of our tool in a real-world case study.

The present paper is structured as follows: Section 2 discusses the academic literature on soft skills; Section 3 describes the materials and methods used to develop SkillNER; Section 4 presents the results produced by the implementation of the NER system; Section 5 demonstrates an application of SkillNER; and Section 6 discusses in greater depth the contribution of this paper, focusing on the possible future developments.

## 2. Background

In this section, we provide an overview of the scientific background on soft skills. In particular, Section 2.1 introduces the taxonomies of skills, namely the main reference sources defined by the international competence frameworks. In Section 2.2, we provide a map of how academic literature has used these sources.

### 2.1. The taxonomies of skills

The taxonomies of skills are dictionaries that classify occupations and skills in different countries. The primary sources of occupational information are ESCO[3] (European) and O*NET[4] (American).

ESCO is a multilingual system that classifies jobs, capabilities, competences, and qualifications relevant to the labour market in Europe. The aim of this framework is to provide an overview of the relationship among skills, profiles, and qualifications in order to fill the gap between academia and industry in Europe. The structure of ESCO is represented in Fig. 1. The occupation classification corresponds to ISCO-O8, which is the *International Standard Classification of Occupations* (International Labour Organization, 2012). One ISCO occupation could correspond to multiple ESCO occupations or to a single one. Each ESCO occupation is characterised by a heterogeneous number of skills (essential or optional). The ESCO structure is based on three pillars (occupations, skills, and qualifications) that are interlinked to show the relationships among them.

O*NET, the American equivalent of ESCO developed for the U.S. Department of Labor, comprises occupations from the Standard Occupational Classification (SOC) system and their corresponding skills, knowledge, and abilities. Each job profile has quantitative information about the level and importance for every owned skill described above. The conceptual foundation of the O*NET framework is represented in Fig. 2, which shows the most important information contained in the database. The main quantitative and qualitative differences between the two taxonomies are shown in Table 3.

Figs. 1 and 2 show that the granularity of the occupations and skills is strikingly different, as can be inferred from the values in Table 1. ESCO has a greater level of detail than O*NET, with six times as many skills and three times as many job profiles. Furthermore, ESCO assigns a large number of different skills to a single job profile, while O*NET has all the descriptors in Fig. 2 assigned to every job profile. Finally, there is no clear distinction between hard and soft skills in O*NET, while around 110 skills are labelled as *transversal*[5] in ESCO (v1.0.3).

These characteristics of ESCO guided the decision to analyse this database as the first case study for SkillNER, as shown in Section 5.

### 2.2. The use of ESCO and O*NET as data sources

Fig. 3 illustrates how previous studies have used ESCO and O*NET. The map should be read as follows: the author "n" addresses the need of firms or institutions, developing a solution ("result") analysing O*NET or ESCO.

It is evident from Fig. 3 that:

- The works having predictive purposes (Frey & Osborne, 2017; Acemoglu et al., 2011) and effective policy design (Alabdulkareem et al., 2018; Autor & Dorn, 2009; MacCrory et al., 2014) are mainly founded on O*NET, possibly because its level of detail is better suited to econometrics studies;
- ESCO is widely used to automatise analysing CVs in the recruitment process (Mirski et al., 2017; Alfonso-Hermelo et al., 2019; Pryima et al., 2018), a possible explanation being the granularity of ESCO skills that makes it easier to match them with CV skills;
- A smaller number of works concern the private sector, probably due to the variability of the data across firms and the reluctance of private sector operators to publish them;
- To the best of our knowledge, no research exists on firms' skill assessment developed through ESCO and O*NET, maybe due to the process being conducted internally and the value obtained not being shared.

To sum up, the study of the literature suggests the lack of a solution which is transversal, thus able to offer an answer to heterogeneous stakeholders with multiple purposes in different sectors. SkillNER is a step in this direction, with a specific focus on soft skills.

---

[2] https://mysterious-hollows-20657.herokuapp.com/

[3] https://ec.europa.eu/esco/portal/home
[4] https://www.onetcenter.org
[5] File transversalSkillCollection.csv

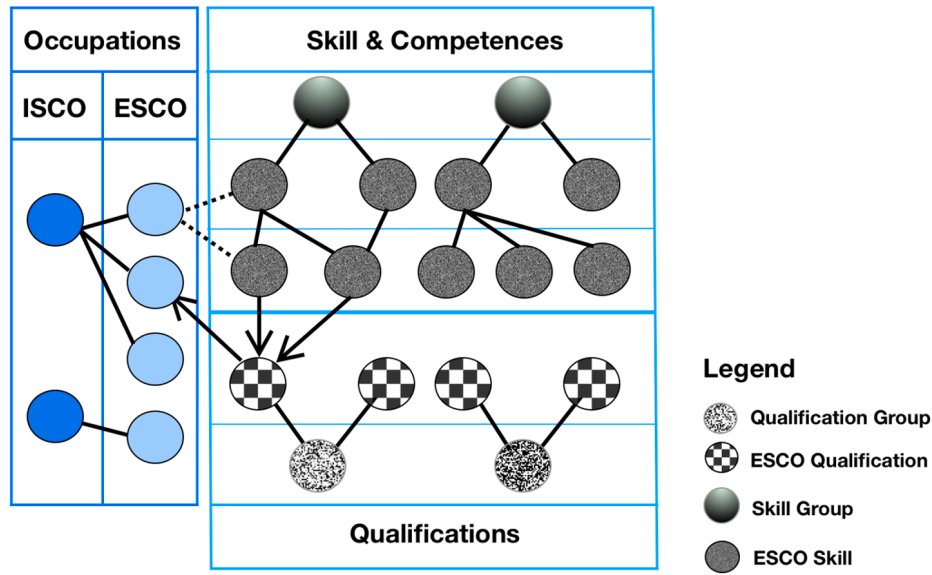

**Fig. 1.** Representation of the ESCO hierarchical structure.

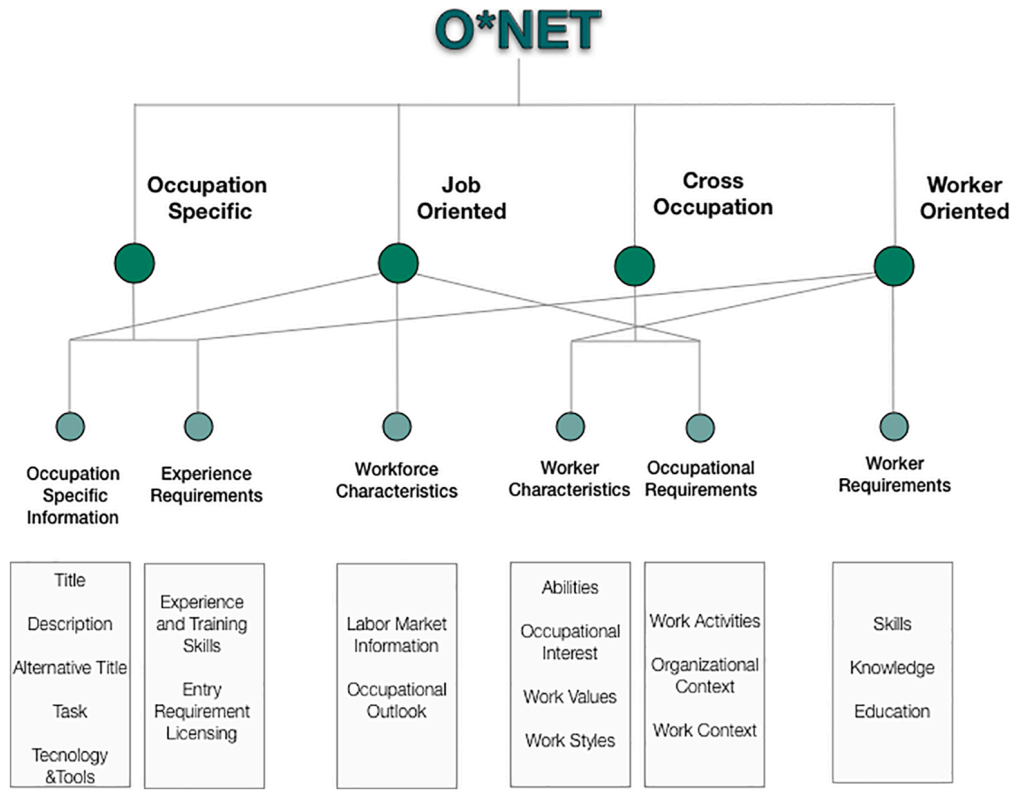

**Fig. 2.** The O*NET content model.
Source: https://www.onetcenter.org/content.html

**Table 1**
The main quantitative (and qualitative) differences between ESCO and O*NET.

| Feature | ESCO | O*NET |
| --- | --- | --- |
| N. of Skills or Descriptors[14] | 13,485 | 277 |
| N. of Occupations | 2942 | 974 |
| Soft Skill | Yes, labelled as "transversal" | No |

[14] Organised into the "content model".

### 3. Materials and methods

This section describes the NER system that we designed to collect soft skills from text.

Recent developments in NER systems demonstrate how the problem of extracting uncommon entities could be approached by focusing on the context (Devlin et al. 2018). For this reason, we describe here the textual

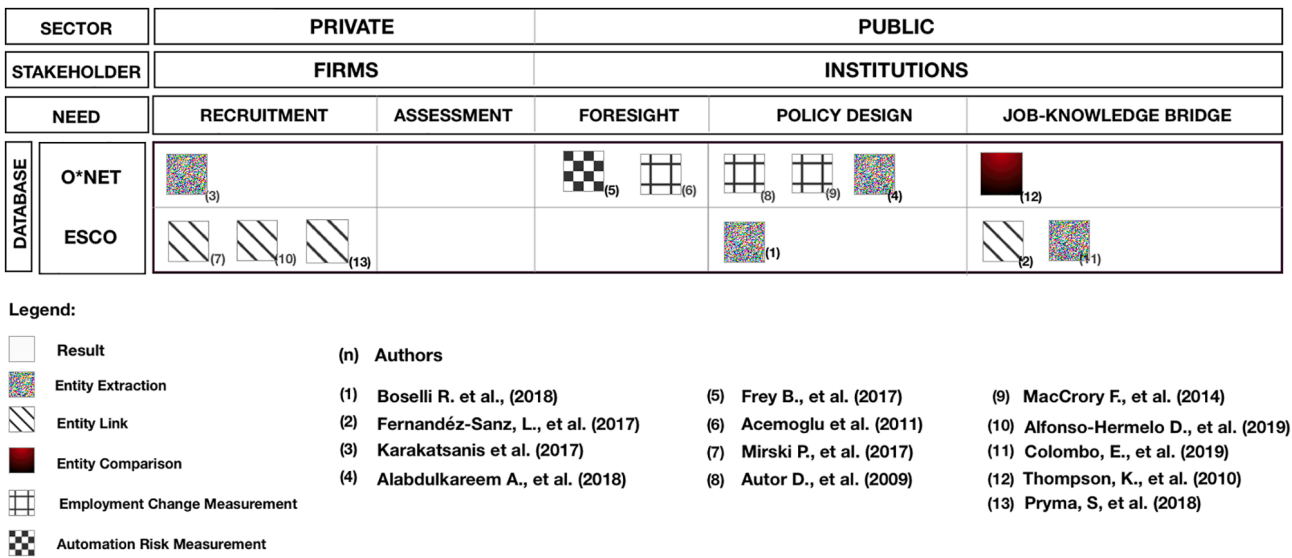

**Fig. 3.** Literature review map, representation by authors.

context in which soft skills appear. Let us consider the example presented in Fig. 4.

Fig. 4 shows two soft skills ("solve difficult problems" and "active listening") surrounded by words that introduce them ("ability to" and "development of"), creating the so-called extraction context. The extraction context is thus made up of two elements that constitute the pillars of SkillNER:

- The *Entity*: a linguistic sign (i.e., one word or a set of words) that references the soft skill;
- The *Clue*: a set of terms, lexical expressions, or recurrent patterns correlated with the appearance of the soft skill.

SkillNER is a supervised NER system whose implementation involved the following phases:

- *Clue Extraction*: in this first phase, we extracted and validated a list of clues.
- *Skill Extraction*: in this second phase, we manually annotated a corpus using as input the validated list of clues.
- *Training and Evaluation:* in this phase, we chose the best supervised model trained on the annotated corpus using different training approaches.

Fig. 5 shows the flow diagram with four different elements graphically displayed: activities (rectangular shape), check points (diamond shape), documents created from the procedure (sheet of paper shape), and databases (database shape).

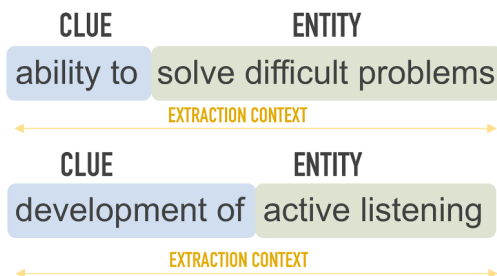

**Fig. 4.** Examples of soft skills within their extraction contexts.

### 3.1. Clue extraction

This phase aimed to extract a list of clues and involved several steps. First, we manually built a seed list of soft skills. In particular, we chose to collect the soft skills referenced in the following sources:

- The three most often cited papers on soft skills according to the Scopus database, with explicit reference to the topic in the title, abstract, and keywords of the papers. The extraction was made in November 2019 with the following query: "TITLE-ABS-KEY ("soft* skill*" OR "social* skill*" OR "commun* skill*" OR "person* skill*" OR "languag* skill*")";
- The O*NET database since it contains occupational definitions and skills relating to the American workforce.

We used O*NET to perform a cross-validation process: identification of the most often cited soft skills in the literature that are simultaneously present in an occupational framework. Moreover, since O*NET labels consist of a maximum of three terms, this step also guarantees concise formulations of skills, which are therefore more easily traceable in the text.

Second, we compiled a list of documents related to soft skills. In particular, we collected the reports of Skills Panorama[6] – an online portal with data on the skill needs of countries, occupations, and sectors across EU member states.

Third, we automatically collected the soft skills extraction contexts. We implemented a *rule-based matcher* in Spacy, a Python built-in NLP tool that allows the identification of specific pieces of text (Honnibal et al., 2017). This system was implemented by defining rules according to specific patterns. In order to understand how patterns work, let us consider the following examples, which are written using the *rule-based matcher*[7] of Spacy:

[{"POS":"NOUN"},{"OP":"*"},{"LEMMA":"solve"}, {"LEMMA":"problem"}]

[{"LEMMA":"solve"},{"LEMMA":"problem"},{"OP":"*"}, {"LEMMA":"soft"},{"LEMMA":"skill"}]

These examples show patterns that extract respectively:

(i) the extraction context that goes from a noun to the words "solve" and "problem";

---
[6] https://skillspanorama.cedefop.europa.eu/en
[7] https://spacy.io/usage/rule-based-matching

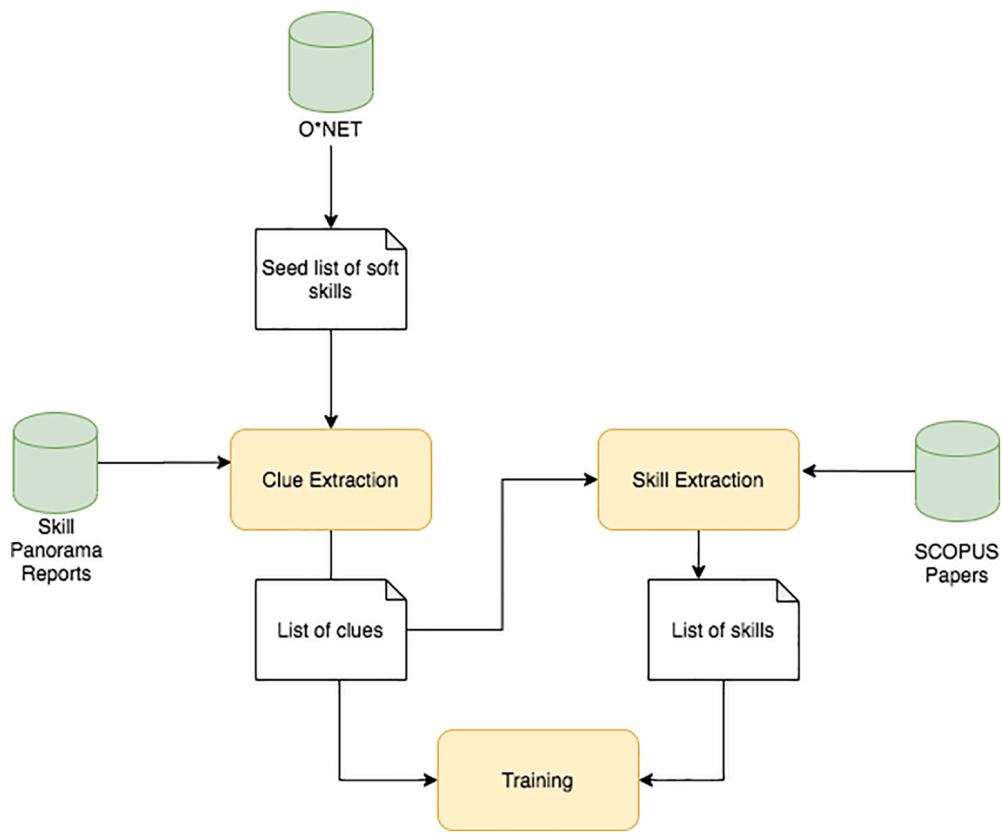

**Fig. 5.** Methodological steps to building SkillNER.

(ii) the extraction context that goes from the words "solve" and "problem" to the words "soft" and "skill".

Finally, we extracted the list of clues from the collected extraction contexts. In this last phase, we focused on keeping only the clues with high frequency. In particular, we followed the Pareto principle: we retained clues that contributed to reaching the 80% of the cumulative frequency of occurrence. The cumulative frequency is calculated taking into account the variants of similar groups of words. For example, in the statements "ability to solve problems" and "ability in problem solving," we consider "ability to" and "ability in" as a unique clue (made up of "ability" and its variants) with a frequency of 2.

*3.2. Skill extraction*

The extraction of skills required three steps. First, we downloaded from Scopus the abstracts of scientific papers where the phrase "soft skills" appears in the title, abstract, or keywords. We choose to rely on scientific production because it is an authoritative source of reliable information vis-à-vis other widely used sources (e.g., Wikipedia, news portals, or social networks).

Second, we searched the list of clues in the scientific corpus. In particular, we implemented a *rule-based matcher* (see Section 3.1. that led us to identify all the sentences in which at least a clue appears.

Finally, we annotated these sentences by involving a group of experts that consisted of four psychologists and four HR specialists. The annotation process was performed in a double-blind mode using the annotation tool called *Prodigy*[8]. Each expert annotated a sample of sentences with the following assignment: "Please select, inside the sentence, that part of text that corresponds to an extraction context of a soft skill." This process led to annotating the corpus with the BIO annotation schema (Ramshaw & Marcus, 1999). The schema is the following:

- B-EXTR: the token is the beginning of an entity representing an extraction context;
- I-EXTR: the token is the continuation of a sequence of tokens representing an extraction context;
- O: for all the other cases.

In practice, it is specified whenever a soft skill appears in the corpus, either alone or together with a clue. It is taken into account that an extraction context can appear with or without a clue. For example, the sentence "The evaluation of participants is based on the assessment of their level of critical thinking and problem solving" would be tagged as follows: "The evaluation of participants is based on the assessment of their < extr > level of critical thinking</extr > and < extr > problem solving</extr > ."

*3.3. Training and evaluation*

The aim of this phase was to train the classification model that incorporates SkillNER. We evaluated two different models and chose the better one in terms of accuracy. In particular, we used the annotated dataset to train two models according to the following learning approaches: (i) feature-based supervised learning and (ii) deep learning.

A SVM is employed for feature-based training. It is one of the supervised machine learning models that produces a linear hyperplane dividing the underlying data either in a positive or a negative category (Vapnik, 2013). SVM is a non-probabilistic classifier that can deal with a large number of features with high accuracy. For this reason, it is well suited to text categorisation problems with no large number of categories to predict, in particular NER (Chiarello et al., 2018). We utilised the LIBSVM library (Chang & Lin, 2011) configured to use a linear kernel with the following linguistic features: lemma, part of speech, and dependency label. These features are extracted from the corpus using the built-in parser of Spacy with the *en_core_web_lg*[9] model that is an English multi-task CNN trained on OntoNotes.

---

[8] https://prodi.gy/

A multilayer perceptron (MLP) is employed for the deep learning training approach. MLP (Schalkoff, 2007) is a class of artificial neural networks that consists of at least three layers of nodes: an input layer, a hidden layer, and an output layer. Each node is a neuron that uses a nonlinear activation function. MLP utilises a supervised learning technique called backpropagation for training. This deep learning approach is also well suited for NER tasks (Gallo et al., 2008). In this study, we used the word embeddings proposed by Mikolov et al. (2013) to represent our inputs into a real-valued vector and designed a sequential MLP model built on the top of the embedding layer, dense layer, and dropout layer. The size of feature vector representation was specified as 50 dimensions, and the number of neurons used in the hidden layer was aligned to 64. The activation function used in the hidden layer and output layer was Relu and Softmax, respectively (Glorot et al., 2011). Moreover, we utilised the Adam optimiser, and the batch size was fixed to 128 (Kingma et al., 2015). The maximum length of the sequence was set to 50, and a total of 20 epochs was employed to train the network model. The configuration of the system resembles that of other works facing a similar task (Chiarello et al., 2018; Speck & Ngomo, 2017; Nguyen & Nguyen, 2017).

The evaluation of the two models was performed by measuring the *precision, recall,* and *f1-score* at a token level. *Precision* is the percentage of named entities found by the learning system that are correct; *recall* is the percentage of named entities present in the corpus that are found by the system; and *f1-score* is the harmonic mean of the precision and recall.

## 4. Results

The initial seed list of soft skills was compiled from the following three papers:

- "*Scaffolding and Achievement in Problem-Based and Inquiry Learning: A Response to Kirschner, Sweller, and Clark*" (Hmelo-Silver et al., 2007)
- "*Graduate employability, 'soft skills' versus 'hard' business knowledge: A European study*" (Andrews & Higson, 2008)
- "*Executive Perceptions of the Top 10 Soft Skills Needed in Today's Workplace*" (Robles, 2012)

These three publications feature respectively 1003, 358, and 385 citations[10] and list a total of 21 different soft skills. However, we only kept the skills that are also mentioned in the O*NET database. Thus, the seed list was composed of the following soft skills: "problem solving," "active reasoning," "communication," "professionalism," "leadership," "teamwork," and "flexibility."

A search for these skills was then conducted in 15 open-access documents from the Skills Panorama dataset provided by CEDEFOP.[11] We collected reports on replacement demand, skills challenges, polarisation of skills, and job growth creators in the European labour market.

Thus, we automatically collected a total of 374 extraction contexts and then obtained a final list of 12 clues by filtering the frequency of occurrences according to the Pareto approach (see Section 3.1). Table 2 shows the top five clues ranked by their aggregated frequencies.

A corpus of 5,359 abstracts gathered from Scopus was used to perform the primary extraction of soft skills. We then collected 850 sentences containing at least one clue from these documents. The manual annotation led us to identify 1,245 extraction contexts (842 not counting the repetitions). Fig. 6 displays an example sentence annotated with three extraction contexts, while Table 3 shows a sample of the extraction contexts containing only the soft skill.

**Table 2**
List of top five clues (words that surround soft skills) appearing in CEDEFOP reports ranked by their frequency of occurrence.

| CLUE | Frequency |
| --- | --- |
| Ability (to/in/of) | 64 |
| Capability (to/in/of) | 35 |
| Level (in/of) | 20 |
| Know-how (in/of) | 19 |
| Proficiency (in/of/at) | 10 |

**Table 3**
Sample of 10 soft skills manually extracted from the corpus of scientific papers.

| Skills |
| --- |
| Empathy |
| Abstract reasoning |
| Address emotions |
| Assertiveness |
| Compassion |
| Conflict management |
| Encode emotions |
| Empowering |
| Non-verbal decoding |
| Manage a good diction |

The annotated input for the two classifiers comprises 2,123 sentences with 15,121 tokens (11,237 not counting the repetitions). Table 4 shows the evaluation metrics in the experimental results. Note that recall is higher than precision in both classifiers. Since SVM yields more reliable results, it is the algorithm incorporated into SkillNER.

A demonstration of SkillNER is made available via a web application[12]. We deployed the soft skill extraction model into a Spacy model and built an application with Streamlit[13] where it is possible to try SkillNER.

## 5. Case Study: Soft skills and job profiles

This section shows an application of SkillNER. We employ the system to discover how soft skills are mentioned in the job profile descriptions provided by ESCO.

ESCO lists a total of 13,485 unique skills for 2,942 unique job profiles. Each skill is represented by a label and then described in natural language, which makes this database suitable for applying SkillNER. In particular, we use our NER system to identify which ESCO skill could be considered "soft." This extraction of soft skills leads us to identify 409 soft skills across 1,243 job profiles.

At this point, we can further explore the results of the extraction by creating two graphs:

- a skill graph Gs = (Vs, Es), where vertices, Vs, are the skills, and edges, Es, are the co-occurrences of the skills in the same job profile. Here, we investigate the existence of clusters of soft skills according to the shared job profiles;
- a job profile graph Gj = (Vj,Ej), where vertices, Vj, are the job profiles, and edges, Ej, are the number of skills the two vertices have in common. In this case, we investigate the existence of clusters of job profiles according to the shared soft skills.

The graphs have been built from the adjacency matrix (N, N), where N is the number of unique skills or job profiles, and the elements in the matrix indicate the number of co-occurrences of the skills in the same job profile. We then use the adjacency matrix to generate an undirected graph of skill Gs = (Vs, Es), where vertices, Vs, are the skills, and edges, Es, are weighted on the co-occurrences of the skills in the same job profile. We do the same for the job profiles graph Gj = (Vj,Ej), where vertices, Vj are the job profiles, and edges, Ej, are the number of skills the two job profiles have in common. In these structures, the higher the weight associated with the edge, the higher the co-occurrence of the skills or job profiles, and the stronger the relationship between them.

---

[9] https://spacy.io/models/en#en_core_web_lg
[10] Latest update from Scopus: 02/15/2021
[11] https://skillspanorama.cedefop.europa.eu/en
[12] https://mysterious-hollows-20657.herokuapp.com/
[13] https://www.streamlit.io/

Fig. 6. Example of annotated sentence from an abstract (Harun & Salamuddin, 2014).

Table 4
Evaluation metrics.

|  | Precision | Recall | F1-score |
| --- | --- | --- | --- |
| SVM | 68.1 | 77.8 | 72.6 |
| MLP | 59.1 | 65.7 | 62.2 |

A network is said to have a community structure if the nodes can be easily grouped into subsets. For this reason, we further explore the structure of these networks by performing a cluster analysis in order to determine the existence of communities of soft skills and profiles.

This cluster analysis is performed by using a weighted modularity approach (Blondel et al., 2008). In general, modularity is an abstract quantity we assign to a partition of the nodes of a network into groups. It is intended to measure how well the partition under consideration represents the natural subdivision of the nodes based on how strongly connected the nodes within each group are (Lambiotte et al., 2019). We choose the subdivision of the nodes that provides the highest modularity and use the resolution limit of modularity (Fortunato & Barthélemy, 2007) as a parameter-dependent approach to investigate how we could best partition the skills and the job profiles. The overall network analysis is performed using the Gephi software (Bastian et al., 2009). Table 5 shows the results of the analysis for graph Gs and Gj. The job graph is more populated since it has ten times as many nodes and six times as many edges as the skill graph. Th job graph also has a higher average in-degree, which signals the thickness of the relationships among nodes.

Figs. 7 and 8 show the two networks using Gephi with the Force Atlas algorithm (Jacomy et al., 2014) for the layout. In this layout, two nodes are represented closely if they share an edge, and the closeness is proportional to edge weight. In this way, nodes that belong to the same communities of nodes (can be grouped into sets so that each set is densely connected internally) but do not share any edge are represented closely. In other words, the visualisations tend to be coherent with the clustering algorithm. The size of the node is proportional to its in-degree, while the colour signifies the cluster to which each node belongs. Finally, only the labels associated with nodes with higher frequency are shown.

The graph of soft skills (Fig. 7) consists of 20 different communities. The composition of the clusters partially confirms what the literature states: it is possible to detect a "leadership" cluster (Winkelmann & Bertling, 2011) which mostly consists of "delegate task," "motivate others," and "persuasion," all of which are characteristics that enable successful workers to interact effectively with others (Bass, 1998).

Table 5
Graph analysis results.

| Graph | Nodes | Edges | Average in-degree |
| --- | --- | --- | --- |
| Gs: Skill graph | 409 | 4336 | 8.54 |
| Gj: Job graph | 1243 | 23,455 | 29.78 |

Cluster 8 embodies the trait of being independent, and it is basically populated by "confidence," "autonomy," and "self-esteem." The "conflict management"(Winkelmann & Bertling, 2011) cluster, surrounded by "empathy" and "being balanced," proves the importance of having good emotion regulation, especially in the workplace (Gross & Thompson, 2007). It is interesting to note that cluster eight is populated by both "emotional intelligence" and "abstract reasoning" – two skills that are usually taught in distinct communities. Moreover, our analysis showed a great number of job profiles containing both "creativity" and "analytical thinking"; the two concepts are frequently considered complementary, and their synergistical presence in multiple job profiles deserves attention and further investigation.

The network graph displayed in Fig. 8 is made up of seven different clusters. Each node is a job profile, and the size of the node gives an indication of the occupations that most require soft skills. A single cluster consists of workers who share similar soft skills. Cluster 0 is populated by managers, job profiles that share communication, planning, and problem-solving skills; Cluster 1 is made up of job profiles belonging to the sphere of law and characterised by persuasion, memory, and the use of a rich vocabulary; Clusters 2 and 4 consist of artistic professions and are linked by creativity, originality, and innovation; Cluster 3 is made up of artisans, who have high precision and meticulousness; Cluster 5 contains medical professionals, who possess a great sense of responsibility, ethics, and critical thinking; finally, Cluster 6 is populated by engineers and architects, whose main characteristics are high precision, the ability to focus, and teamwork.

Further discussion of the graph and the policy-related consequences of our analysis results are outside the scope of the present paper. We summarise the main applications of the results in the final section.

## 6. Conclusions and future developments

In this paper, we introduce a methodology to automatically extract soft skills from text. The solution that we present is called SkillNER, a supervised NER system trained on a scientific corpus annotated by a panel of experts. To promote the use of our approach by other scholars in studying soft skills, SkillNER is also made available as a web application[14]. This paper also shows a preliminary application of the system by extracting soft skills from the job profile descriptions provided by ESCO. This application leads to detecting communities of job profiles and communities of soft skills, which are discovered by analysing two networks built considering job profiles that share soft skills and soft skills that share job profiles.

Our results provide a methodological step forward that can open a more quantitative discussion on the role of soft skills in the labour market. We summarise the contribution of this paper to the labour market in Table 6.

In conclusion, our work has certain limitations. First, from a computational point of view, SkillNER could be improved in terms of accuracy. The introduction of recent deep neural architectures (such as transformers) is driving dramatical improvements in NLP tasks. BERT (Devlin et al., 2018), for example, is a language model establishing new standards for NER. What positions BERT as the best model for language learning is the fact that it is trained in a particular neural architecture (called transformer) that is able to learn a specific task with scant labelled data. We are aware that the use of such a language model would improve the accuracy of SkillNER. However, BERT should be trained on a specific corpus to improve the effectiveness of the NER. Training it for soft skills identification was outside the scope of this paper. Second, the training process could be improved. As the interest in soft skills grows, the volume of textual data increases. This leads to more data available for labelling and feeding into the training process. The wider the training corpus, the greater the accuracy of any NLP system. This would be the case for SkillNER.

---

[14] https://mysterious-hollows-20657.herokuapp.com/

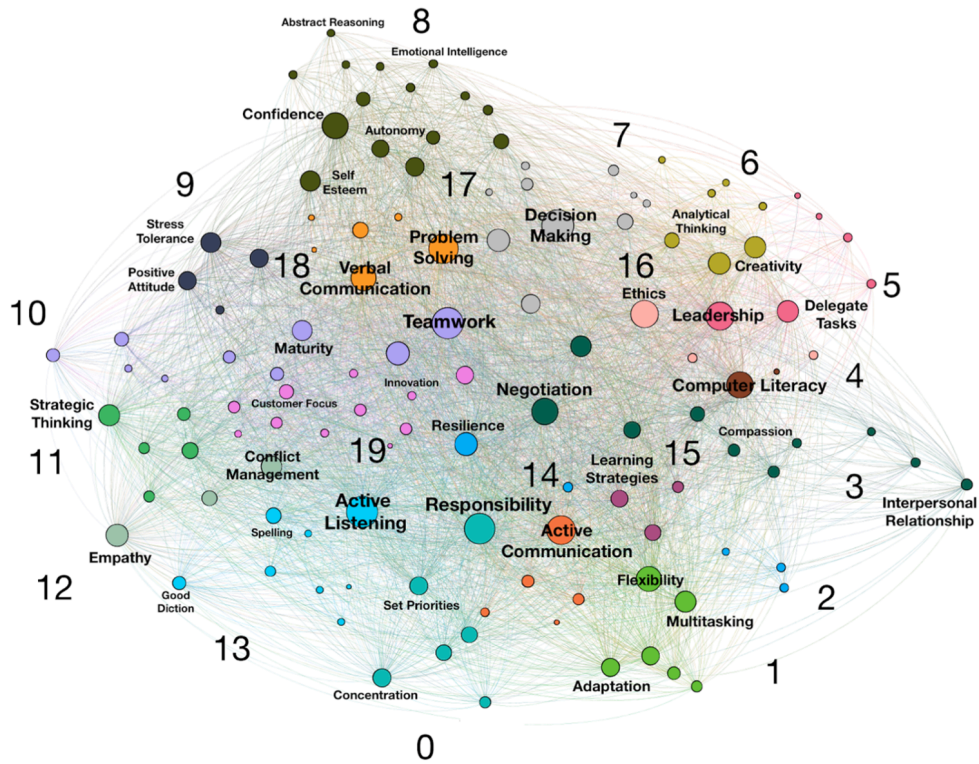

**Fig. 7.** A network graph representation of the soft skills extracted and the clusters in which they are found (only the most relevant nodes are shown).

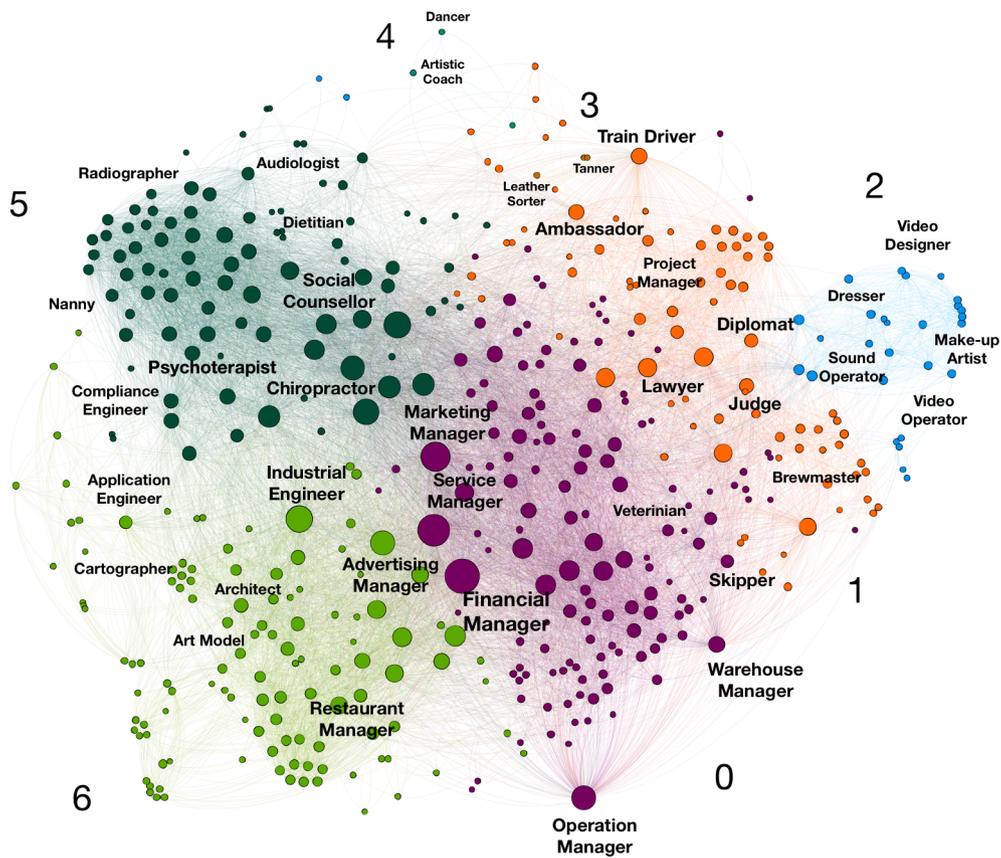

**Fig. 8.** A network graph representation of the job profiles and the clusters in which they are found (only the most relevant nodes are shown).

**Table 6**
A summary table presenting the three outputs of the research, the stakeholders potentially interested in them, the final purpose of the outputs, and the activities that could be performed with them.

| Output | Stakeholder | Purpose | Activity |
| --- | --- | --- | --- |
| *Skill(N)ER* | Firms | Recruitment | CV analysis |
| | | Assessment | Identification of soft skill lacks |
| | | | Ad hoc learning paths development to foster portability of soft skills |
| | Institutions | Updating international skills databases | Build data-driven soft skills ontology |
| *Communities of Soft Skills* | Firms | Updating job descriptions | Enhance the number of soft skills following the relations visualised through the graph |
| | Institutions | Updating international skills satabases | |
| | Workers | Skill portability | Identifying the skills more easily acquirable in relation to the ones currently possessed |
| *Communities of Job Profiles* | Firms | Job-knowledgebridge | Developing common soft skills courses for different job profiles according to the relations visualised through the graph |
| | Institutions | | Customise university courses |
| | | Foresight | Providing evidence of the most resilient job profiles |
| | | Policy design | |
| | Workers | Job portability | Identifying the nearest job profile in relation to the one currently performed |

Three key avenues should be explored in future research. First, as mentioned above, a transformer-based architecture and a language model could be used to train the supervised model. Second, it would be possible to use word embeddings or domain-independent knowledge bases (such as WordNet or ConceptNet) to explore the semantic similarity among the skills and then find clusters of soft skills based on their meaning. Third, it would be necessary to test the method on different domains, exploring additional textual data sources (job descriptions, CVs, and patents). This would improve the potentialities of SkillNER in bringing order to an otherwise confusing conceptual landscape.

**CRediT authorship contribution statement**

**S.a. Fareri:** Writing - original draft, Conceptualization, Visualization, Project administration. **N.b. Melluso:** Writing - original draft, Methodology, Software, Formal analysis. **F.c. Chiarello:** Writing - review & editing. **G.d. Fantoni:** Conceptualization, Supervision.

**Declaration of Competing Interest**

The authors declare that they have no known competing financial interests or personal relationships that could have appeared to influence the work reported in this paper.

*Acknowledgments*

This study was partly founded by the EU project ULISSE (Understanding, Learning and Improving Soft Skills for Employability) Call 2018 – KA203 - Erasmus+ "Strategic Partnerships for Higher Education" Project ID: 2018-1-IT02-KA203-048286.

The authors would like to thank Erre Quadro srl for the precious help in the analysis, the Career Office of Unipi and in particular dr. Antonella Magliocchi for her continuous support.

**Appendix A. Supplementary data**

Supplementary data to this article can be found online at https://doi.org/10.1016/j.eswa.2021.115544.